\documentclass[conf]{new-aiaa}

\usepackage[dvipsnames]{xcolor}

\usepackage[utf8]{inputenc}

\usepackage{graphicx}
\usepackage{amsmath}
\usepackage[version=4]{mhchem}
\usepackage{siunitx}
\usepackage{longtable}

\definecolor{tablegreen}{RGB}{55, 126, 36}

\title{Human-Computer Interaction Glow Up: Examining Operational Trust and Intention Towards Mars Autonomous Systems}

\author{Thomas Chan\footnote{Assistant Professor, Positive Augmented Research and Development, California State University Northridge}, Jeremy Argueta, Jazlyn Armendariz, Allison Graham, Sarah Hwang}
\affil{Autonomy Research Center for STEAHM, California State University Northridge}

\author{Basak Ramaswamy\footnote{Product Owner, Mission Planning, Sequencing and Analysis, Jet Propulsion Laboratory, California Institute of Technology}, So Young Kim\footnote{Manager, Human Systems Integration, Jet Propulsion Laboratory, California Institute of Technology}, \& Scott Davidoff\footnote{Manager, Human-Centered Design, Jet Propulsion Laboratory, California Institute of Technology}}
\affil{Jet Propulsion Laboratory, California Institute of Technology}

\begin{document}

\maketitle
\begin{abstract}
Tactful coordination on earth between hundreds of operators from diverse disciplines and backgrounds is needed to ensure that Martian rovers have a high likelihood of achieving their science goals while enduring the harsh environment of the red planet. The operations team includes many individuals, each with independent and overlapping objectives, working to decide what to execute on the Mars surface during the next planning period. The team must work together to understand each other's objectives and constraints within a fixed time period, often requiring frequent revision. 

This study examines the challenges faced during Mars surface operations, from high-level science objectives to formulating a valid, safe, and optimal activity plan that is ready to be radiated to the rover. Through this examination, we aim to illuminate how planning intent can be formulated and effectively communicated to future spacecrafts that will become more and more autonomous.

Our findings reveal the intricate nature of human-to-human interactions that require a large array of soft skills and core competencies to communicate concurrently with science and engineering teams during plan formulation. Additionally, our findings exposed significant challenges in eliciting planning intent from operators, which will intensify in the future, as operators on the ground asynchronously co-operate the rover with the on board autonomy. Building a marvellous robot and landing it onto the Mars surface are remarkable feats – however, ensuring that scientists can get the best out of the mission is an ongoing challenge and will not cease to be a difficult task with increased autonomy.

\end{abstract}



\section{Introduction}
\lettrine{O}perating a rover on Mars to advance our knowledge about the universe's formation and future requires deep expertise in many scientific and engineering domains. Scientists who study the geology and atmosphere of Mars must negotiate with scientists seeking traces of ancient life to decide the most valuable data we can collect at any point\cite{vertesi2015seeing, vertesi2020shaping}. 

In addition, the scientists have to work with instrument engineers who know how to utilize onboard instruments best to collect the data as efficiently and accurately as possible. All of these instruments have to negotiate the resource usage checking in with vehicle system engineers. Vehicle system engineers, which are further divided into sub-systems, are responsible for enabling observations, returning data to earth, and maintaining the overall safety of the rover. In the end, a fully detailed and validated activity plan is uplinked to the rover to execute. Under normal conditions, execution only minimally deviates from the plan\cite{vertesi2015seeing, vertesi2020shaping}. 

Future of space operations will involve more autonomous spacecraft where some of the decisions currently made on the ground will be delegated to the flight system. The autonomous flight system has to work with the operators on the ground to achieve their intent, which requires operators to communicate their intent to the flight system. The key motivation for this study was to observe -- how human operators communicate with each other during the planning process to inform how planning intent could be formulated for the more autonomous spacecrafts of the future. 

We conducted a series of semi-structured interviews with science and engineering operations representatives from two Mars surface missions to answer this question. We focused on how operators exchange information regarding available resources, constraints, and planning intent, how input from one team was translated into actions across the other team. We analyzed the interview recordings by delving into how decisions were made, what kind of language was used, and what range of soft skills were crucial. An atomic research model \cite{pidcock2018atomic} a common technique in user experience research, was utilized to generate and test emergent mechanistic themes linked to the planning and negotiation process. Results found that three latent themes were most commonly linked to foster mission coordination:

1.	Corporate Knowledge Gluers – people who fill in any mission-relevant knowledge gaps to update rotating or missing personnel;
2.	Bridge Builders – people who are conduits to foster collective understanding between science and engineering teams;
3.	Efficiency Optimizers – people who help streamline the tasks and attention of teams to meet mission deadlines. 

Not only were these skills necessary for smooth and efficient planning operations, these three main latent themes were associated with specific operations roles which were created months after the operations began, as the need for better coordination and communication became apparent. Furthermore, we observed that planning intent is gradually formulated through many iterations and interactions between science and engineering teams. 

The end goal of this work is to inform the design of autonomous planning systems which can allow users to express their intent seamlessly and co-operate a rover with trust.  Our study provides a human centered approach to this collaboration problem, in contrast with research in autonomy domain, which focuses on how to ensure spacecraft can achieve intent once it is perfectly formulated. 

In the next section we give an overview of Mars surface operations to provide readers some context about our study. In section III, we describe the study protocol and data analysis methods employed. Results section provides a more granular description of socio mechanical themes summarized above. In the following section we discuss additional observations and several areas of further investigation to enable a happy collaboration between humans and autonomy.





\section {Mars Surface Operations in a Nutshell}
In this section we provide a high level overview of Mars surface operations. Note that operations are very intricate, covers a long duration, includes multiple processes with different life cycles and involves many roles. Operators need to train for months and shadow senior operators to be able to partake. Our description below is a brief summary to help readers interpret our study results in context. 

The goal of any space mission is to achieve high level science goals that are conceptualized years before operations begin. Scientists start with \textit{strategic plans} development by setting deadlines for milestones to meet in the next 2 to 3 month time frame. The strategic plans are produced taking into account overall capabilities of the rover, such as how far it can drive in a single day, or how many observations can be conducted in average in a day. 

A \textit{long term plan} is derived from the strategic plan, covering the next 3 to 5 sols, looking ahead from the current sol. When the long term plan is developed, scientists have more context to decide on specific science observations, and a more accurate estimate of resources that will be available \cite{erickson2002mars}.

Only during tactical planning an exact activity plan is formulated. The process starts with the n+1 portion of the long term plan and the latest data received from the spacecraft describing its most recent status. At the end of the process, every activity has to be defined precisely and translated into commands to be executed in the next plan cycle on Mars, which is often covers one sol (Mars solar day) \cite{mishkin2007human}.  

Broadly speaking, during the tactical process, scientists have to decide which exact observations to perform, while engineering teams inform the scientists about which observations are feasible and how many of them can be accommodated by running various simulations. High level abstractions of activities (e.g. "Take a picture of that rock from this angle") and simulation results are displayed on visual timelines \cite{chung2012timeline} for the team to evaluate options. The operations team  need to come to a consensus of what science activities are of priority for the next day given various constraints - environmental, engineering, and science at the end of the process.

Once an activity plan that is safe for the rover, and one that can achieve optimal science is produced, engineering teams translate the plan into commands that the spacecraft understands. The command products are then uplinked to the spacecraft to execute onboard for the next planning cycle.

\section{Method}
\subsection{Procedures}
A total of 10.0019 hours of semi-structured interviews were conducted to address our targeted inquiries about the Mars surface missions' planning and negotiation processes. The interview questions aimed to elicit both broad and detailed responses regarding planning and negotiations while offering the researchers opportunities to get clarification and gain context for each response.  Questions were piloted with operations personnel and tailored to match commensurate Mars missions experience. Specifically, the decision tree of questions corresponded with the interviewees' experience on \textbf{Mars Science Laboratory-Curiosity Rover (MSL; for details visit https://mars.nasa.gov/msl/home/)} and/or \textbf{Mars2020-Perseverance Rover (M2020; for details visit https://mars.nasa.gov/mars2020/)}.  Recruitment was conducted through email invitations. Invitations were sent to identified vital personnel currently active on either MSL or/and M2020 missions. Interviews were performed virtually using WebEx (Cisco Systems, Milpitas, CA) and lasted between 45-90 minutes, and were transcribed.

\subsection{Participants}
A total of ten semi-structured interviews were conducted with Mars surface mission operators. The roles of participants spanned multiple scientific and engineering capacities. Transparently, although official roles (i.e., titles) were initially targeted to gain the spectrum of planning and negotiating experiences -- most participants noted that they performed multiple roles at different time frames or in parallel. Additionally, their roles were modified, expanded, or promoted to meet emergent mission needs. For instance, Project Scientists also stepped in as Science Operations Coordinators/Science Engineering Liaisons when needed. Similarly, few Science Planners that we interviewed later performed the Tactical Uplink Lead role. Consequently, with much overlap, seven out of ten participants had coordination roles, five out of ten participants focused on science objectives, and five out of ten participants focused on engineering tasks. Nine interviewees were from the JPL organization, while one was outside of JPL. Four participants worked only on the MSL mission, while six were on both MSL and M2020 missions. Seven of the ten participants were classified as "veterans" on Mars missions due to their extensive work history. The other three participants were considered in the "early stages" of their mission operations career.   

\subsection{Analysis Plan}
Interviews were transcribed, reviewed by each team member, and then moments extracted and coded by team members into an Atomic Research\cite{sharon2016validating, pidcock2018atomic} repository. The coding of key observations was guided by grounded theory—an inductive approach that utilizes constant comparative methods to establish analytic distinctions across multiple iterations \cite{charmaz2020pursuit}. Emergent themes were developed from interrelated categories to better characterize and connect multiple observations. Following the process of atomic user research, facts from the observations were anonymized and codified from video records in an Airtable repository \cite{airtable}. Emergent patterns were rigorously tested for their generalizability (i.e., applied to other interviewees' experiences). If generalizability was achieved, more than three common experiences, then they were labeled as insights and aggregated into the higher-level categories that we report on in our findings section below.

\section{Results}

\begin{figure}[hbt!]
\centering
\includegraphics[width=0.75\textwidth]{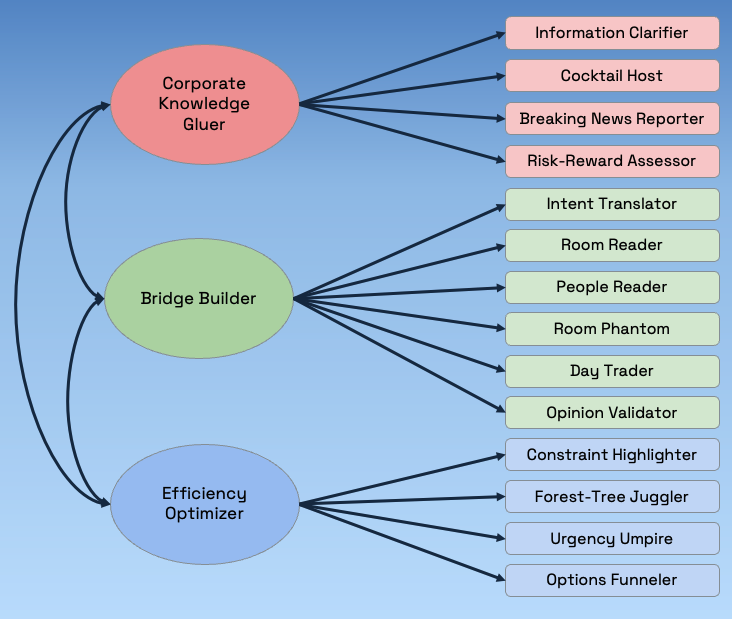}
\caption{Overview of the Latent Themes}
\label{fig:RO}
\end{figure}

Our analyses of semi-structured interviews yielded three emergent latent variable themes with 14 specific functions, as shown in Fig. \ref{fig:RO}. Latent themes are the overarching [unobserved] constructs derived from a composite of the observed measures -- in our case, the actual functions performed \cite{muthen2009statistical}. Functions were the observed units generated from systematic analyses and honed operationalizations by the current study's researchers. Again, latent themes and functions were only identified if they were applied across at least three negotiation and planning mission contexts. 

\subsection{Latent Theme I - Corporate Knowledge Gluer - 28.09\% of all total quotes}
This first latent theme illustrates the critical role of Mars mission people who fill in any mission-relevant knowledge gaps to update rotating or missing personnel. Corporate Knowledge Gluers perform functions that focus on maintaining the continuity of mission awareness -- reducing gaps in critical knowledge, and provide updates to lessen learning curves. An exemplar quote that highlights this high-level role is \textit{"...the main part of that role is just coordinating and keeping everybody up to date about what everybody else is doing...each role can get very focused on their little piece of the puzzle...basically that person putting all the pieces together and making sure they all fit."} -P2Q19.  The continuity Corporate Knowledge Gluers provide are highlighted even more during off-nominal conditions. They "get everyone up to speed," and need to do it quickly to ensure that contingency plans are negotiated in time to make adjustments. The specific functions that make up the Corporate Knowledge Gluer composite: 
\begin{enumerate}
\item \textit {Information Clarifier - Strengthens understanding of objectives and clears up any misunderstandings during negotiations.}  This function performed by Corporate Knowledge Gluers targets ensuring that current or changing objectives are understood by science, engineering, and operations teams. The people from this function draws upon what has happened during the mission and explains the current course. An exemplary quote that illustrates this function, "You might hear two requests that sounded slightly contradictory...bringing those forward to the rest of the team...and clarifying anything that was unclear" -P1Q2.

\item \textit {Cocktail Host - Acts as the day to day point of contact to maintain mission structure.}  This function targets the stable presence of the "go-to person" who knows Sol's current course and objectives. Cocktail Hosts also direct any person who needs more detailed updates to connect. An exemplary quote that showcases this function, "“That person who’s...listening to every conversation...efficiently make sure that the right people are hearing the right information and understand the constraints and the intent of why we're doing things" -P5Q4.

\item \textit {Breaking News Reporter - Broadcasts updates when plans go off-nominal.} This function performed focuses on gather and disseminating critical updates as mission objectives go off course. Breaking News Reporters have their eyes and ears on all aspects of the day's current plans and use their daily connections with all teams to get and transmit updates in a speedy matter. An exemplary quote that highlights this function, "If there's some late-breaking knowledge of vehicle state...late-breaking news is, "Oh, hey, we can't do that...we don't have enough time...let's just go on to the thing that we want to do strategically" -P10Q17. 

\item \textit {Risk-Reward Assessor - Explains the costs associated with a given option.} This function performed targets transmitting mission knowledge of the risks associated with given decisions. Risk-Reward Assessors do not make specific decisions, but they process the costs and benefits of each potential decision at the highest levels. An exemplary quote that illustrates this function, "We think about what's really most important... what's worth the cost that the engineering folks have now placed on the site from a resource or time perspective" -P4Q8.

\end{enumerate}

\subsection{Latent Theme II - Bridge Builder - 40.45\% of all total quotes}
This second latent theme illustrates the vital role of Mars mission persons in being conduits to foster collective understanding between science and engineering. Bridge Builders use "soft skills" such as translating both verbal and nonverbal languages to fill in gaps for a diverse set of science and engineering teams. The exemplary quote that of this high-level theme is,\textit{"In the morning, I work with the scientists and understand what their goals for the day are...I then transfer and keep communicating with the various engineering roles...of facilitating for that to happen"} -P2Q17. Science and engineers, in general, have different priorities, terminologies, and focuses – these Bridge Builders offer ago-between to ensure that information amongst the teams does not "get lost in translation." Especially involving the expression of scientific intentions and protection of the equipment – this function becomes critical in maintaining and improving rapport between people who may have different directions during mission planning and negotiations. The specific functions that make up the Bridge Builder composite:

\begin{enumerate}

\item \textit {Intent Translator - Communicates an improved comprehension of planned goals between scientists and engineers.} This function performed uses "social intelligence" to match the amount of contextual information needed to link to what is behind the scientific intentions. Once the context is given, another level of interpreting the necessary steps to engineers is given to best direct them to fulfill the intended activities. An exemplary quote that highlights this function, "We struggle with the science to engineering communication...I put a lot of thought into...how to translate the science into intent and then ask that the engineers understand" -P4Q27.

\item \textit {Room Reader- Senses the overall tone of the team as situations unfold.} This function enables Mars personnel's social intelligence systems to gauge the present mission milieu. Room Readers are especially attuned to gauge the non-verbal cues during off-nominal conditions when team members may become overwhelmed -- and this is conveyed to other members to match tasks accordingly.  An exemplary quote of this function, “We often talk about getting the temperature of the room...if everybody's stressed out...I better find out more about that because this can affect planning today" -P3Q5.

\item \textit {People Reader - Gauges specific personalities and working styles of individuals.}  This function also uses people's social intelligence systems to target conveying the apt interactions style and corresponding information based on individuals' characteristics. An exemplary quote of this function, “A lot of my role is...reading the person I'm talking to...Jane really likes making the decisions on her own. So I just give Jane the facts...John, you can be like, “I think that this is the right thing to do" -P5Q37.

\item \textit {Room Phantom - Toggles between rooms of different teams and listens for important updates.} This function employs their executive set-shifting abilities as Room Phantoms are engaged with multiple conversations and appear when relevant updates occur. These people vacillate between being passive or active in conversations -- depending on the priority of the information. An exemplary quote of this function, "I run between the rooms during operations...during remote world, I've got four WebExs at once... I'm talking to four groups at once and I'm listening" -P5Q36.

\item \textit {Day Trader - Facilitates compromises from the negotiations to achieve consensus.} This function enables people to become a broker of mission objectives exchanges to ensure planning decisions converge with maximizing buy-in from personnel. An exemplary quote of this function, "Anticipate the arguments that people are going to have and propose compromises early...everyone felt like they were getting a little something" -P9Q1.

\item \textit {Opinion Validator - Acknowledges individuals' viewpoints and voices to increase their acceptability with consensus decisions.} This function ensures people's contributions to mission planning and negotiations are heard. Mainly during periods under time crunches, not everyone's opinions could be thoroughly discussed. Opinion Validators become the buffer to prevent people from being disgruntled or disengaged in the future in the planning and negotiation process. An exemplary quote of this function, "The purpose of planning is to make it so that everybody has had a voice, everybody's been able to state their opinion...so that the plan that comes out, everybody's like...I'm good with that" -P8Q11.

\end{enumerate}

\subsection{Latent Theme III - Efficiency Optimizer - 31.46\% of all total quotes}
This third latent theme illustrates the essential role of Mars mission persons to streamline the tasks and attention of teams to meet mission deadlines. Efficiency Optimizers process all inputs at the meta-level and direct focus on the top priorities, and ensure those priorities do not get lost -- especially under mission time constraints to communicate with Martian rovers. An exemplary quote that illustrates this high-level theme is, \textit{“Show people clearly what the constraints are and...how best we can address their desires within the constraints. The sooner you do that, the sooner you avoid a discussion that is unproductive”} -P9Q9. These "make it happen" people go beyond just stating the apparent objectives by identifying what is pressing both in the short- and long-term objectives, and process the planning and barriers to make those objectives successful. The specific functions of the Efficiency Optimizer composite: 
\begin{enumerate}
\item \textit {Constraints Highlighter - Emphasizes the restrictions associated with a given planning activity.} This function spotlights the limitations of the resources being used to conduct particular science activities. Constraints Highlighters constantly communicate with engineering and science teams to ensure that plans are possible given the available resources. Exemplary quote for this function, “If we're using too much power, then I let the [anonymized] lead know that something has to go" -P6Q8.

\item \textit {Forest-Tree Juggler - Zooms in and out between long-term blueprints and current situations.} This function uses people's ability to go between big picture scientific objectives with the current plan being implemented. Forest-Tree Jugglers keep track of the everyday decisions and whether they have implications or potentially jeopardize significant mission objectives. Exemplary quote for this function, “How should we prioritize the mission time, over months and years? Where should we drive next?...that's kind of the strategic planning work that I help facilitate” -P3Q18.

\item \textit {Urgency Umpire - Facilitates teams to focus on immediate mission needs.} This function brings forth the planning and decisions that need to be made rapidly to achieve mission objectives. Urgency Umpires push teams to focus on particular objectives and relegate other tasks to the background until completion.  Exemplary quote for this function, “We'll say...this observation is really high priority...because we're going to need this to make a decision tomorrow morning" -P7Q10.

\item \textit {Options Funneler - Narrows down decisions for science and engineering teams to pick from during planning and negotiations.} This function presents a truncated set of options for teams to make their decisions on. Options Funnelers use their informed judgment to ensure that the most viable decisions are filtered and presented.  Exemplary quote for this function, “You are a facilitator, you are not...the decision maker...you provide people with a set of choices" -P2Q8.

\end{enumerate}

\begin{longtable}[l]{|p{0.2\textwidth}|p{0.1\textwidth}|p{0.25\textwidth}|p{0.35\textwidth}|}

\caption{\label{tab:table-1}Codes generated to describe identified functions. Abbreviations are used for three latent categories Corporate Knowledge Gluer (CKG), Bridge Builder (BB), Efficiency Optimizer (EP)} \\

\hline
Functions              & Percentage of Latent Composition & Description of theme                                                                            & Examples from the open ended responses                                                                                                                                                                                \\ \hline
\textcolor{Salmon}{\textbf{CGK:}} Information Clarifier  & 20.83\%                & Strengthens understanding of objectives and clears up any misunderstandings during negotiations & “You might hear two requests that sounded slightly contradictory…bringing those forward...and clarifying anything that was unclear."                                                                                  \\ \hline
\textcolor{Salmon}{\textbf{CGK:}} Cocktail Host          & 20.83\%                & Acts as the day to day point of contact to maintain mission structure                           & “That person who’s...listening to every conversation...efficiently make sure that the right people are hearing the right information and understand the.constraints and the intent of why we're doing things.”        \\ \hline

\textcolor{Salmon}{\textbf{CGK:}} Breaking News Reporter & 16.67\%                & Broadcasts updates when plans go off-nominal                                                    & ‘ “...if there's some late breaking knowledge of vehicle state...late breaking news is, "Oh, hey, we can't do that...we don't have enough time...let's just go on to the thing that we want to do strategically”.’    \\ \hline

\textcolor{Salmon}{\textbf{CGK:}} Risk-Reward Assessor   & 20.83\%                & Explains the costs associated with a given option                                               & “We think about what's really most important...what's worth the cost that the engineering folks have now placed on the site from a kind of resource or time perspective.”                                             \\ \hline
\textcolor{tablegreen}{\textbf{BB:}} Intent Translator      & 16.67\%                & Communicates an improved comprehension of planned goals between scientists and engineers        & “We struggle with the science to engineering communication...I put a lot of thought into...how to translate the science into intent and then ask that the engineers understand.”                                      \\ \hline
\textcolor{tablegreen}{\textbf{BB:}} Room Reader            & 13.89\%                & Senses the overall tone of the team as situations unfold                                        & “...we often talk about getting the temperature of the room...if everybody's stressed out...I better find out more about that because this can affect planning today.”                                                \\ \hline
\textcolor{tablegreen}{\textbf{BB:}} People Reader          & 16.67\%                & Gauges specific personalities and working styles of individuals                                 & ‘ “A lot of my role is...reading the person I'm talking to...Jane really likes making the decisions on her own. So I just give Jane the facts...John, you can be like, “I think that this is the right thing to do”.’ \\ \hline

\textcolor{tablegreen}{\textbf{BB:}} Room Phantom           & 11.12\%                & Toggles between rooms of different teams and listens for important updates                      & “I run between the rooms during operations...during remote world, I've got four WebExs at once...I'm talking to four groups at once and I'm listening."                                                               \\ \hline
\textcolor{tablegreen}{\textbf{BB:}} Day Trader             & 13.89\%                & Facilitates compromises from the negotiations to achieve consensus                              & “Anticipate the arguments that people are going to have and propose compromises early...everyone felt like they were getting a little something."                                                                     \\ \hline
\textcolor{tablegreen}{\textbf{BB:}} Opinion Validator      & 13.89\%                & Validates individuals' opinions to increase their acceptability with consensus decisions        & “...the purpose of {[}anonymized{]} planning is to make it so that everybody has had a voice, everybody's been able to state their opinion...so that the plan that comes out, everybody's like...I'm good with that.” \\ \hline

\textcolor{CornflowerBlue}{\textbf{EP:}} Constraint Highlighter & 28.57\%                & Emphasizes the constraints of a given planning activity                                         & “...if we're using too much power, then I let the {[}anonymized{]} lead know that something has to go.”                                                                                                               \\ \hline
\textcolor{CornflowerBlue}{\textbf{EP:}} Forest-Tree Juggler    & 17.86\%                & Zooms in and out between long-term blueprints and current situations                            & “How should we prioritize the mission time, over months and years? Where should we drive next?...that's kind of the strategic planning work that I help facilitate.”                                                  \\ \hline
\textcolor{CornflowerBlue}{\textbf{EP:}} Urgency Umpire         & 10.71\%                & Facilitates teams to focus on immediate mission needs                                           & “...we'll say...this observation is really high priority...because we're going to need this to make a decision tomorrow morning."                                                                                     \\ \hline
\textcolor{CornflowerBlue}{\textbf{EP:}} Options Funneler       & 21.43\%                & Narrows down decisions for science and engineering teams to pick from                           & “You are a facilitator, you are not...the decision maker...you provide people with a set of choices."      \\ \hline

\end{longtable}

\section{Discussion and Future Work}

Using our observations, we observe that activity planning for a spacecraft can be summarized as optimizing the return of high-value and -volume science data, with resource availability (e.g. memory, energy), and spacecraft health and safety. We observed many individuals from multiple disciplines participating in this optimization calculus in a kind of scientific congress. The coordination and communication challenge amongst the operators gave rise to the creation of key "linchpin" roles to ensure a smooth and efficient planning process. We argue that these roles serve two primary purposes.

First, people occupying these roles need to be able to speak the language of different domains, such that they can succinctly express the \textbf{constraints} that must be adhered to and how these constraints affect the \textbf{option space}. They also have to explain the long- and short-term \textbf{implications} of picking one option over the other to the various science team members. And second, because the calculus of negotiation can create lead to friction that undermines team morale, our findings call attention to the fact that constraints are actualized through the expert practice of "soft skills" functions, We highlight the key Bridge Builder functions that facilitate the human-relational factors linked to mission success -- and overall longevity of organizational and team trust forging \cite{siau2018building}. Specifically, Bridge Builders were the most cited functions identified when discussing the planning and negotiation process. Bridge Builders were commonly cited as the medium through which scientific intent and communications between engineering and science teams were honed -- exemplified by the quote, "Why did we do the thing that we did yesterday? from...an engineering perspective, but also from the science perspective...I really do walk that line, I have to understand both the engineering side and the science rationale." Beyond these Bridge Builders carrying the corporate knowledge -- they ensure the creation, maintenance, and evolution of the social milieu \cite{zak2017trust} -- i.e., MSL as a "Well Oiled Machine." The Well Oiled Machine that is the MSL mission is primarily due to the scaffolding of science and engineering team members. That is, the processes and workflows for the MSL mission were built from the ground up with science and engineering teams co-creating its inputs, functions, and evolution (e.g., creation of roles). Future promising work may want to focus on the intensive shadowing process beyond what any training manual or algorithm may be able to specify to understand how added new operators or machines enter, learn, and evolve the co-creation of mission-related cultures. 

These findings suggest significant considerations for the design of autonomous systems. Note that when we have more autonomy onboard the spacecraft, we will likely need more sophisticated software tools that analyze likely outcomes and constraints, along with specialized roles to coordinate between science and autonomy, analogous to \textit{Science Operations Coordinator} who currently coordinates between science and engineering domains. Operators' trust will increase if they interact with a system that can analyze risks accurately, communicate them clearly for the information consumer, and recognize the larger mission context. Designers of autonomous systems should consider how corporate knowledge should be maintained, how bridges could be built across domains, and how operators can be guided towards optimal solutions during operations. Even though we may rely on a human versed in the workings of autonomy, the software we use should simplify their job to avoid rendering them as a single point of failure.  

Additionally, we observed that planning intent is elaborated gradually and through many iterations. Scientists really begin the planning process with at best rough estimates of possible science opportunities and resources. As more data is received, and more evaluations have been performed on the ground, more precise constraints and opportunities emerge, and a more complete set of the space of achievable plans comes into view.
These often lead to negotiations among the operators and result in adjustments to the observation plan.

This notion of planning as \textbf{emergent} countervenes somewhat a common assumption we find in the literature surrounding autonomous planning and scheduling systems, which posit that intent can be perfectly formulated a priori. Alternatively, at the very least, we argue that the comprehensive and coherent expression of intent does not entirely lie within the research domain of AI or autonomy developers. Our observations instead reveal challenges faced when eliciting intent from planners while lacking definitive data to guide their decision-making. In the current operations paradigm that we observed, decisions required to more completely specify science intent are elaborated on the ground by enacting iterative comparison of possible futures. In this way, scientists come to incrementally understand their own science intent through conversation with other scientists, that get more and more precise over time. Towards that end, the idea that AI researchers might rely on the perfect expression of science intent might inject an insincere or even impossible precondition on the ultimate infusion of their own research systems to teams in the wild. We would, instead, advocate for an approach where autonomous systems might instead ask for a science intent that is more realistically the outcome of an iterative process.

As operators partially delegate decision-making to the spacecraft, there will be less information on the state of spacecraft during plan execution. Operators will need to consider multiple scenarios where resources and other conditions can vary — eliciting such information from operators comprehensively is not a trivial task — where many permutations must be considered. We advocate that, to more accurately design operations concepts, multi-disciplinary research teams are needed to investigate how software support tools can guide operators to formulate intent coherently, comprehensively so that optimal science outcomes can be achieved. We see this as holding the key to the successful operations of the autonomous missions of the future.

\section{Conclusion}
Infusing autonomy to space missions and their operations requires a great deal of information exchange between the domain experts and the autonomous system. In addition, this information exchange has to be asynchronous by virtue of space operations, where getting real-time confirmation from the earth is impossible due to communication limitations. Such use of autonomy is more ambitious than how standard autonomous systems work today where either humans can take control over anytime (e.g., autonomous cars), or humans are expected to provide confirmation (e.g., medical diagnostic application), or penalty for a mistake is low (e.g., Roomba). 

By studying how this information is exchanged among human operators, we revealed vital issues that need to be addressed while exchanging similar information between autonomy and human operators. We also demonstrated how a user-centered approach could be taken while developing autonomous systems that collaborate with humans. Our investigation signals significant challenges that will be faced and point out essential investigation areas for human-machine teaming to glow up to its full potential.

\section*{Acknowledgments}
This research was carried out in part at the Jet Propulsion Laboratory, California Institute of Technology, under a contract with the National Aeronautics and Space Administration (80NM0018D0004) and Cooperative Agreement (80NSSC19M0200). Funding from National Institutes of Health Bridges to the Doctorate Research Training Program (1T32GM137863) was also provided to support student work on this project. The authors wish to thank the Mars Science Laboratory and Perseverance Rover Mission Operations teams for their generous support. 

\bibliography{main}

\end{document}